\documentclass{llncs}
\usepackage{llncsdoc}
\usepackage{graphicx}
\usepackage{color}
\usepackage{caption}
\usepackage{amssymb}
\usepackage{amsfonts}
\usepackage{amsmath}
\usepackage{dsfont}
\usepackage{multirow}
\usepackage{array}
\usepackage{arydshln}
\usepackage{boldline}
\usepackage{tabu}
\usepackage{wrapfig}
\usepackage{url}

\newcommand*\samethanks[1][\value{footnote}]{\footnotemark[#1]}

\begin{document}
%%%%%%%%%%%%%%%%%%%%%%%%%%%%%%%%%%%%%%%%%%%%%%%%%%%%%%%%%%%%%%%%%%%%%%%%%%%%%%%%%%%%%%%%%%%%%%%%%%%%
\title{Discriminative Localization in CNNs for Weakly-\\Supervised Segmentation of Pulmonary Nodules}
\author{Xinyang Feng\thanks{Both authors contributed equally to this work.}\inst{1} \and Jie Yang\samethanks[1] \inst{1} \and Andrew F. Laine\inst{1} \and Elsa D. Angelini\inst{1,2} }
% index{Feng,Xinyang}
% index{Yang,Jie}
% index{Laine,Andrew}
% index{Angelini,Elsa}
\institute{Dept. of Biomedical Engineering, Columbia University, New York, NY, USA
\and NIHR Imperial BRC, ITMAT Data Science Group, Imperial College, London, UK}

\maketitle
%%%%%%%%%%%%%%%%%%%%%%%%%%%%%%%%%%%%%%%%%%%%%%%%%%%%%%%%%%%%%%%%%%%%%%%%%%%%%%%%%%%%%%%%%%%%%%%%%%%%
\begin{abstract}

Automated detection and segmentation of pulmonary nodules on lung computed tomography (CT) scans can facilitate early lung cancer diagnosis. Existing supervised approaches for automated nodule segmentation on CT scans require voxel-based annotations for training, which are labor- and time-consuming to obtain. In this work, we propose a weakly-supervised method that generates accurate voxel-level nodule segmentation trained with image-level labels only. By adapting a convolutional neural network (CNN) trained for image classification, our proposed method learns discriminative regions from the activation maps of convolution units at different scales, and identifies the true nodule location with a novel candidate-screening framework. Experimental results on the public LIDC-IDRI dataset demonstrate that, our weakly-supervised nodule segmentation framework achieves competitive performance compared to a fully-supervised CNN-based segmentation method. 

\end{abstract}
%%%%%%%%%%%%%%%%%%%%%%%%%%%%%%%%%%%%%%%%%%%%%%%%%%%%%%%%%%%%%%%%%%%%%%%%%%%%%%%%%%%%%%%%%%%%%%%%%%%%

\section{Introduction}

Lung cancer is a major cause of cancer-related deaths worldwide. Pulmonary nodules refer to a range of lung abnormalities that are visible on lung computed tomography (CT) scans as roughly round opacities, and have been regarded as crucial indicators of primary lung cancers \cite{nodule_2005}. The detection and segmentation of pulmonary nodules in lung CT scans can facilitate early lung cancer diagnosis, timely surgical intervention and thus increase survival rate \cite{cancer_early}. 

Automated detection systems that locate and segment nodules of various sizes can assist radiologists in cancer malignancy diagnosis. Existing supervised approaches for automated nodule segmentation require voxel-level annotations for training, which are labor-intensive and time-consuming to obtain. Alternatively, image-level labels, such as a binary label indicating the presence of nodules, can be obtained more efficiently. Recent work \cite{FA,spie} studied nodule segmentation using weakly labeled data without dense voxel-level annotations. Their methods, however, still rely on user inputs for additional information such as exact nodule location and estimated nodule size during the segmentation.

Convolutional neural networks (CNNs) have been widely used for supervised image classification and segmentation tasks. It was very recently discovered in a study \cite{cam_2016} on natural images that CNNs trained on semantic labels for image classification task (``what''), have remarkable capability in identifying the discriminative regions (``where'') when combined with a global average pooling (GAP) operation. This method utilizes the up-sampled weighted activation maps from the last convolutional layer in a CNN. It demonstrated the localization capability of CNNs for detecting relatively large-sized targets within image, which is not the general scenario in medical imaging domain where pathological changes are more various in size and rather subtle to capture. However, this work sheds light on weakly-supervised disease detection.
%which can greatly alleviating the labeling task required for training.

In this work, we exploit CNN for accurate and fully-automated segmentation of nodules in a weakly-supervised manner with binary slice-level labels only. Specifically, we adapt a classic image classification CNN model to detect slices with nodule, and simultaneously learn the discriminative regions from the activation maps of convolution units at different scales for coarse segmentation. We then introduce a candidate-screening framework utilizing the same network to generate accurate localization and segmentation. Experimental results on the public LIDC-IDRI dataset \cite{lidc_2011,tcia_2013} demonstrate that, despite the largely reduced amount of annotations required for training,  our weakly-supervised nodule segmentation framework achieves competitive performance compared to a CNN-based fully-supervised segmentation method. 

%%%%%%%%%%%%%%%%%%%%%%%%%%%%%%%%%%%%%%%%%%%%%%%%%%%%%%%%%%%%%%%%%%%%%%%%%%%%%%%%%%%%%%%%%%%%%%%%%%%%
\section{Method}
\begin{figure}[t]
\centering
\includegraphics[trim={0 0 0 1mm},clip,width=12cm]{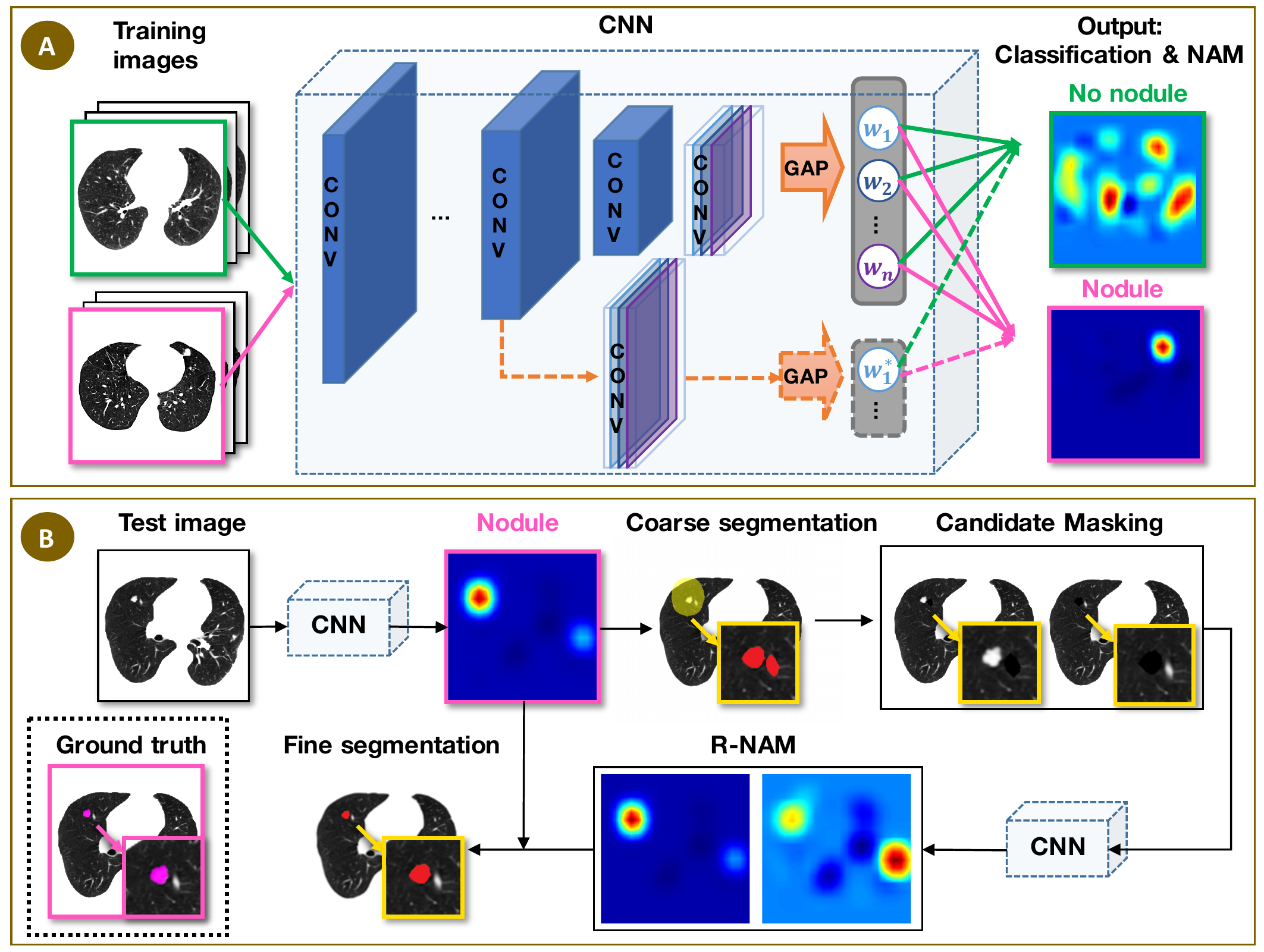}
\caption[ ] {(A) Training: a CNN model is trained to classify CT slices and generate nodule activation maps (NAMs); (B) Segmentation: for test slices classified as ``nodule slice'', nodule candidates are screened using a spatial scope defined by the NAM for coarse segmentation. Residual NAMs (R-NAMs) are generated from images with masked nodule candidates for fine segmentation.}
\label{Fig:framework}
\end{figure}

The framework is overviewed in Fig. \ref{Fig:framework}. There are two stages: training stage and segmentation stage. In the first stage, we train a CNN model to classify CT slices as with or without nodule. The CNN is composed of a fully convolutional component, a convolutional layer + global average pooling layer (Conv+GAP) structure, and a final fully-connected (FC) layer. Besides providing a binary classification, the CNN generates a nodule activation map (NAM) showing potential nodule localizations, using a weighted average of the activation maps with the weights learnt in the FC layer. In the second stage, coarse segmentation of nodule candidates is generated within a spatial scope defined by the NAM. For fine segmentation, each nodule candidate is masked out from the image alternately. By feeding the masked image into the same network, a residual NAM (called R-NAM) is generated and used to select the true nodule. Shallower layers in the CNN can be concatenated into the classification task through skip architecture and Conv+GAP structure, extending the one-GAP CNN model to multi-GAP CNN that is able to generate NAMs with higher resolution.

\setlength{\textfloatsep}{15pt}

\subsection{Nodule Activation Map}

In a classification-oriented CNN, while the shallower layers represent general appearance information, the deep layers encode discriminative information that is specific to the classification task. Benefiting from the convolutional structure, spatial information can be retained in the activations of convolutional units. Activation maps of deep convolutional layers, therefore, enable discriminative spatial localization of the class of interest. In our case, we locate nodules with a specially generated weighted activation map called nodule activation map.

\subsubsection{One-GAP CNN} For a given image $I$, we represent the activation of unit $k$ at spatial location $(x,y)$ in the last convolutional layer as $a_k(x,y)$. The activation of each unit $k$ is summarized through a spatially global average pooling operation as $A_k=\sum_{(x,y)} a_k(x,y)$. The feature vector constituted of $A_k$ is followed by a FC layer, which generates the nodule classification score (i.e. input to the softmax function for nodule class) as: 

\begin{equation}
S_\mathrm{nodule} = \sum\nolimits_k w_{k,\mathrm{nodule}} A_k=\sum\nolimits_k w_{k,\mathrm{nodule}} \sum\nolimits_{(x,y)} a_k(x,y)
\end{equation}
where the weights $w_{k,\mathrm{nodule}}$ learnt in the FC layer essentially measure the importance of unit $k$ in the classification task. As spatial information is retained in the activation maps through $a_k(x,y)$, a weighted average of the activation maps results in a robust nodule activation map: 

\begin{equation}
\mathrm{NAM}(x,y) = \sum\nolimits_k w_{k,\mathrm{nodule}} a_k(x,y)
\end{equation}
% maybe normalize a_k across the batch, since the absolute value of a_k affects the classification score
The nodule classification score can be directly linked with the NAM by:
\begin{equation}
S_\mathrm{nodule} = \sum\nolimits_{(x,y)} \sum\nolimits_k w_{k,\mathrm{nodule}} a_k(x,y) = \sum\nolimits_{(x,y)} \mathrm{NAM}(x,y)
\end{equation}

By simply up-sampling the NAM to the size of the input image \textit{I}, we can identify the discriminative image region that is most relevant to nodule.
\subsubsection{Multi-GAP CNN} Although activation maps of the last convolutional layer carry most discriminative information, they are usually greatly down-sampled from the original image resolution due to pooling operations. We hereby introduce a multi-GAP CNN model that takes advantage of shallower layers with higher spatial resolution. Similar to the idea of the skip architecture proposed in fully-convolutional network (FCN) \cite{fcn_2016}, shallower layers can be directed to the final classification task skipping the following layers. We also add a Conv+GAP structure following the shallow layers. The concatenation of feature vectors generated by each GAP layer is fed into the final FC layer. The NAM generated from the multi-GAP CNN model (multi-GAP NAM) is a weighted activation map involving activations at multiple scales.

\subsection{Segmentation}
\subsubsection{Coarse Segmentation}
For slices classified as ``nodule slice", nodule candidates are screened within a spatial scope $C$ defined by the most prominent blob in the NAM processed via watershed. They are then coarsely segmented using an iterated conditional mode (ICM) based multi-phase segmentation method \cite{icm}, with the phase number equal to four as determined by global intensity distribution.
\subsubsection{Fine Segmentation} The NAM indicates a potential but not exact nodule location. To identify the true nodule from the coarse segmentation results, i.e. which nodule candidate triggered the activation, we generate residual NAMs (R-NAMs) by masking each nodule candidate $R_j$ alternately and feeding the masked image $I \backslash R_j$ into the same network. Significant change of activations within $C$ indicates the exclusion of a true nodule. Formally, we generate the fine segmentation by selecting the nodule candidate $R_k$ following:
\begin{equation}
\displaystyle R_k=\mathrm{argmax}_{R_j}\ \sum\nolimits_{(x,y) \in C} \big[\mathrm{NAM}_I(x,y)-\mathrm{NAM}_{I \backslash R_j}(x,y)\big]^2
\end{equation}
where $\mathrm{NAM}_I$ is the original NAM, and $\mathrm{NAM}_{I \backslash R_j}$ is the R-NAM generated by masking nodule candidate $R_j$. Our current implementation targets the segmentation of one nodule per NAM. Incidence of slices with two nodules is $\sim1\%$ within slices with nodules. No slices contain more than two nodules in our dataset.
\subsubsection{Multi-GAP Segmentation}
For the multi-GAP CNN model, we observed a slight drop in classification accuracy compared with the one-GAP CNN model (see Section \ref{network}), which is expected since features from shallower layers are more general and less discriminative. In light of this, we further propose a multi-GAP segmentation method by training both a one-GAP CNN model and a multi-GAP CNN model to combine the discriminative capability of the one-GAP system and finer localization of the multi-GAP system. 

\setlength{\intextsep}{0mm}
\begin{wrapfigure}{R}{0.57\textwidth}
	\centering
\includegraphics[trim={0 0mm 0 0mm},clip,width=7.25cm]{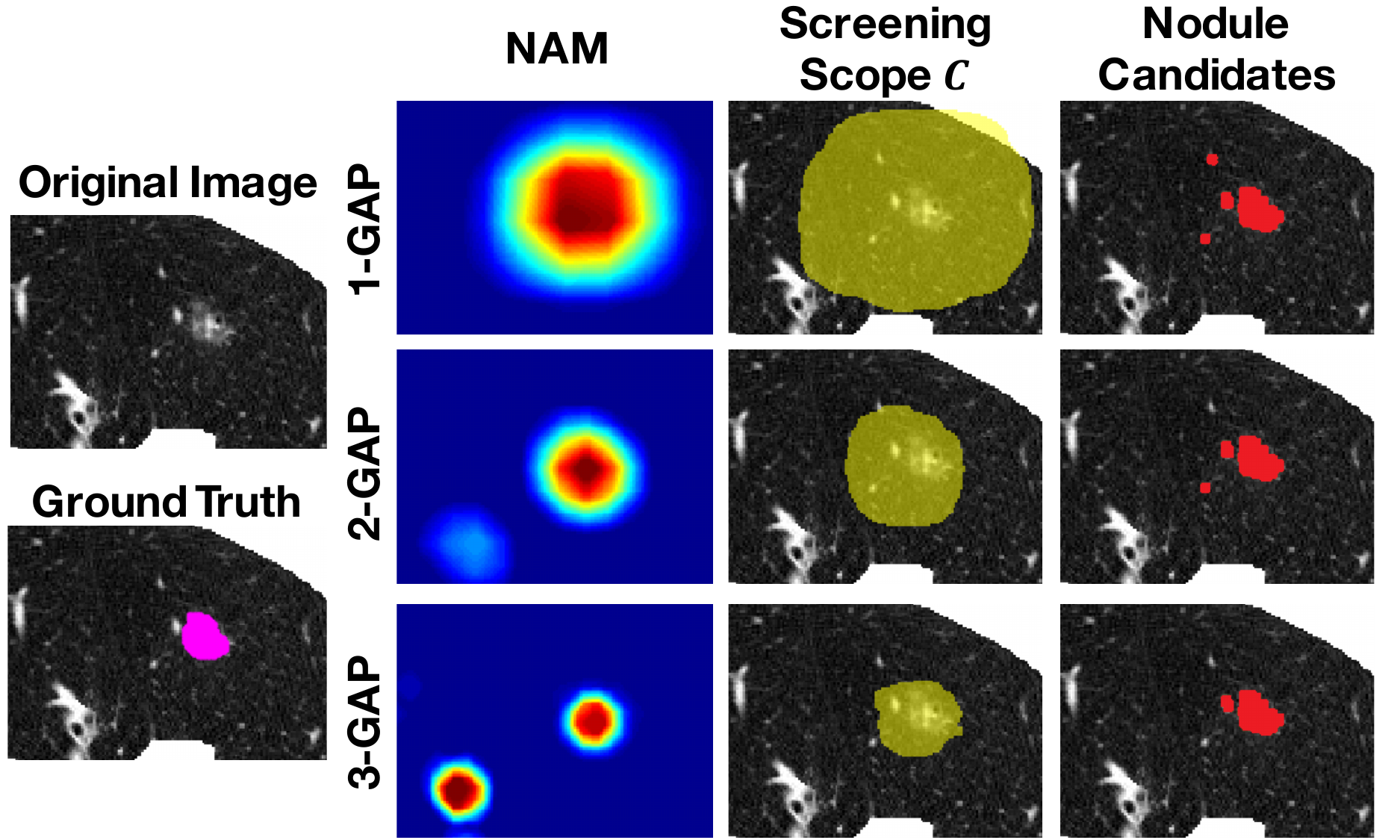}
\caption[ ] {\small{Illustration of 1-/2-/3-GAP NAMs, the screening scopes \textit{C} and coarse segmentation results on a sample slice.}}
%  (architectures of 1-/2-/3-GAP CNNs are described in Section \ref{network})
\label{Fig:gap_illustration}
\end{wrapfigure}

Specifically, segmentation is performed on slices classified as ``nodule slice" by the one-GAP CNN model for its higher classification accuracy. To define the screening scope for coarse segmentation, we first use the one-GAP NAM to generate a baseline scope $C_1$. If there is a prominent blob $C_\mathrm{multi}$ within $C_1$ in the multi-GAP NAM, we define the final scope $C$ as $C_\mathrm{multi}$ to eliminate redundant nodule candidates with more localized spatial constraints. When the multi-GAP NAM fails to identify any discriminative regions within $C_1$, the final screening scope $C$ remains $C_1$. The R-NAM of the masked image is generated by the one-GAP CNN model and compared with one-GAP NAM within $C_1$. Fig. \ref{Fig:gap_illustration} illustrates 1-/2-/3-GAP NAMs, the corresponding screening scopes \textit{C} and coarse segmentation results on a sample slice. While multi-GAP NAM enables finer localization, one-GAP NAM has better discriminative power.

%%%%%%%%%%%%%%%%%%%%%%%%%%%%%%%%%%%%%%%%%%%%%%%%%%%%%%%%%%%%%%%%%%%%%%%%%%%%%%%%%%%%%%%%%%%%%%%%%%%%

\section{Experimental Results}

\subsection{Data and Experimental Setup}

Data used in this study contains 1,010 thoracic CT scans from the public LIDC-IDRI database. Details about this database, such as acquisition protocols and quality evaluations, can be found in \cite{lidc_2011}. Lungs were segmented and each axial slice was cropped to 384\textsf{x}384 pixels centering on the lung mask. Nodules were delineated by up to four experts. Voxel-level annotations are used to generate slice-level labels, and are used as ground truth for segmentation evaluation. Nodules with diameter $<$3mm are excluded \cite{luna16}. Given the high false positive rate of nodule detection, we select slices with nodule if there were overlapped annotations by at least two experts, and select slices without nodule if no expert reported a nodule in the slice. Annotations from different experts were merged using the STAPLE algorithm \cite{staple}. A total of $N_{\mathrm{slice}}=8,345$ slices with nodule are selected, and an equal number of slices without nodule are randomly extracted. The total number of voxels belonging to nodule is $N_{\mathrm{voxel}}=1,658,981$. Segmentation evaluation is focused on slices with one nodule. Rare cases of slices with two nodules are discussed in the end of Section \ref{network}. Training, validation and test sets are generated by distributing the full set of subjects in a ratio of 4:1:1 through stratified sampling so that they have non-overlapping subjects and similar distribution of nodule occurrence.

% The subjects are first stratified, based on the number of nodules in the CT scans, into ten sub-groups.

\begin{table}[t]
\caption{Comparison of Segmentation Performance}
\centering
\begin{tabular}{l|cccccc}
 \hline
Method		 		 & TPR 			& FPR	 		& $\mathrm{FPR_{nodule}}$	& Dice 					& TP Dice 			& TP DOA (mm$^2$) \\
					&				&				&						&mean $\pm$ SD			&mean $\pm$ SD		&mean $\pm$ SD			\\
\hlineB{2}
1-GAP Coarse\  \ \	 	&  \textbf{0.77}*\ \ \ 	& \textbf{0.11}*\textdagger\ \ 	&  \textbf{0.08}*\ \ \ 	&  0.46 \scriptsize{($\pm$0.31)}\ \				& 0.61 \scriptsize{($\pm$0.20)}\ \  		& 57.6 \scriptsize{($\pm$71.1)}\ \ \\ \hline
2-GAP Coarse\  \ \ 		&  0.76\ \ \ \		& -		 		&  0.09 \ \ \		&  0.50 \scriptsize{($\pm$0.34)}\ \ 				& 0.66 \scriptsize{($\pm$0.18)}\ \  		& 41.6 \scriptsize{($\pm$53.6)}\ \ \\ \hline
3-GAP Coarse\  \ \ 		&  0.75\ \ \ \  		& -		 		&  0.11\ \ \ \		&  0.50 \scriptsize{($\pm$0.32)}\ \ 				& 0.67 \scriptsize{($\pm$0.18)}\ \  		& 40.1 \scriptsize{($\pm$50.9)}\ \ \\ \hline
1-GAP Fine   		 	&  0.75\ \ \ \  		& -		 		&  0.11\ \ \ \		&  0.54 \scriptsize{($\pm$0.34)}\ \ 				& 0.73 \scriptsize{($\pm$0.15)}\ \  		& 30.7 \scriptsize{($\pm$52.8)}\ \ \\ \hline
2-GAP Fine   		 	&  0.75\ \ \ \  		& - 	 			&  0.11\ \ \ \		&  0.55*\scriptsize{($\pm$0.33)}\ \ 				& 0.74*\scriptsize{($\pm$0.14)}\ \  		& 29.2*\scriptsize{($\pm$46.8)}\ \ \\ \hline
3-GAP Fine    			&  0.74\ \ \ \ 		& - 	 			&  0.12\ \ \ \		&  0.54 \scriptsize{($\pm$0.34)}\ \ 				& 0.74 \scriptsize{($\pm$0.14)}\ \  		& 29.3 \scriptsize{($\pm$46.4)}\ \ \\ \hline
U-net 		 		&  0.74\ \ \ \ 		& 0.29\ \ \  		&  0.26\ \ \ \		&  \textbf{0.56} \scriptsize{($\pm$0.38)}\ \  		& \textbf{0.76} \scriptsize{($\pm$0.19)}\ \  & \textbf{28.3}\  \scriptsize{($\pm$44.8)}\ \ \\ \hline
\end{tabular}

\scriptsize{* = best performance within our framework; \textbf{boldfaced} = overall best performance;\\
\textdagger= 1-GAP model is used for nodule slice-level detection within our framework.}
\label{Table:performance}
\end{table}

\subsection{Segmentation Performance}\label{network}
\begin{figure}[t]
\centering
\includegraphics[trim={0 2mm 0 2mm},clip,width=11.5cm]{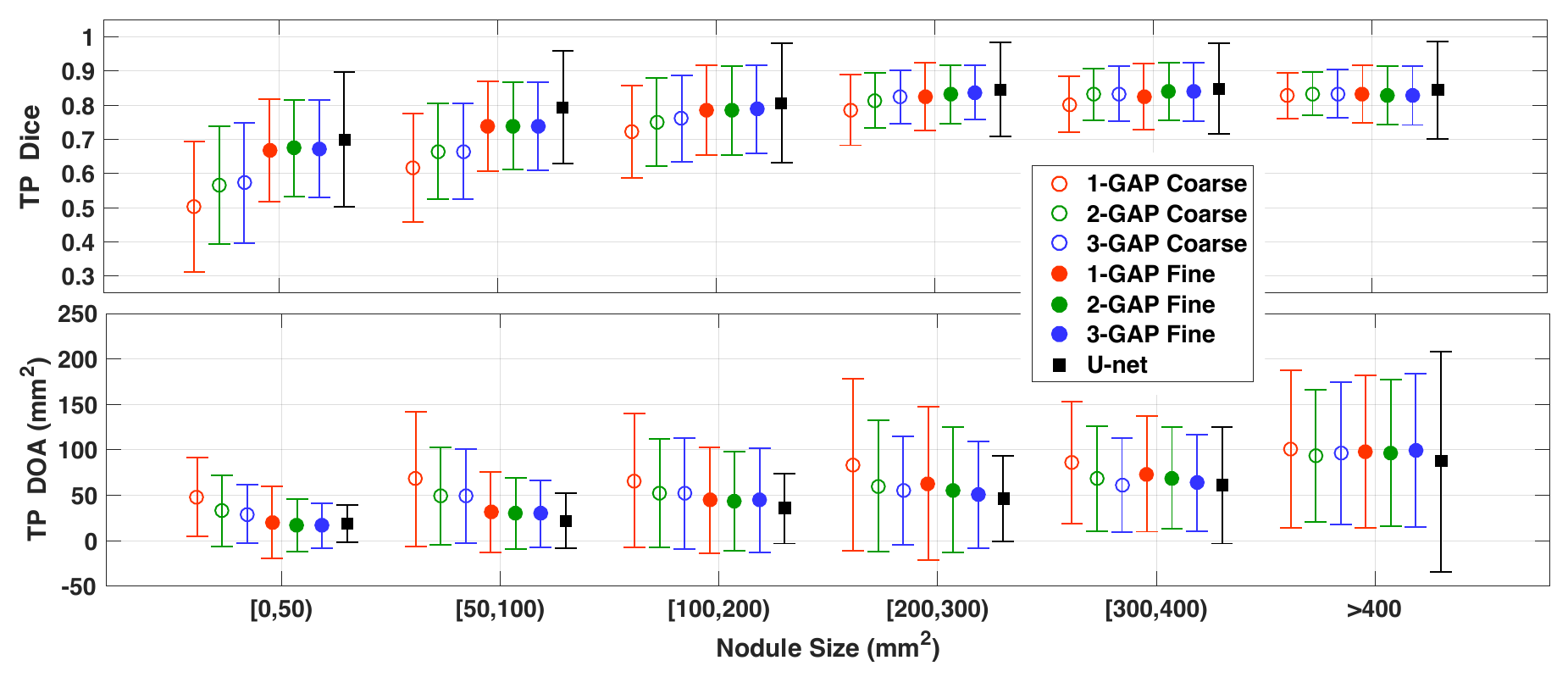}
\caption[ ] {TP Dice and TP DOA  (mean and standard deviation) versus nodule size.}
\label{Fig:performance}
\end{figure}

We compare our framework with a fully-supervised CNN-based method (see below). True positive rate (TPR) of nodule detection, false positive rate (FPR) of ``nodule" detected on slices \textit{without} nodule, false positive rate ($\mathrm{FPR_{nodule}}$) of ``nodule" detected on slices \textit{with} nodule, Dice overlap of nodule segmentation over all slices with nodule (Dice), Dice over truly detected nodules (TP Dice) and absolute difference of segmented areas over truly detected nodules (TP DOA) are reported in Table \ref{Table:performance}. Furthermore, TP Dice and TP DOA versus nodule size are reported in Fig. \ref{Fig:performance}.
\subsubsection{Weakly-Supervised Segmentation based on NAM:} Our network is based on VGG16Net architecture \cite{vgg}, implemented in TensorFlow. The last pooling layer \url{pool5} and the FC layers \url{fc6}, \url{fc7}, \url{fc8} are removed \cite{cam_2016}. The weights of remaining VGG16Net layers are initialized based on the model pre-trained on ImageNet. The Conv+GAP structure is added after \url{conv5_3} layer for 1-GAP CNN, added after \url{conv5_3} and \url{conv4_3} layers for 2-GAP CNN, and added after \url{conv5_3}, \url{conv4_3}, and \url{conv3_3} layers for 3-GAP CNN. The learning rate of the newly added layers is 10 times the learning rate of the remaining VGG16Net layers. We trained using stochastic gradient descent with momentum. The initial learning rate ($10^{-2}$ for 1-GAP, $2\times10^{-3}$ for 2-GAP, $10^{-3}$ for 3-GAP), learning decay (0.99), batch size (30) were set by grid search based on classification accuracy on the validation set. The best accuracy values are 88.4\% for 1-GAP CNN, 86.6\% for 2-GAP model, and 84.4\% for 3-GAP model on the test set.
\subsubsection{Comparison with Fully-Supervised Segmentation:} An adapted model based on U-net architecture \cite{u-net} is used as a fully-supervised CNN-based model for comparison. The cost function is the negative mean Dice coefficient across mini-batch. 
% The same training set of slices with nodule is used to train the model. The trained model is applied to the same set of validation and test slices with nodule. 
The algorithm was optimized with Adam method. The initial learning rate ($2\times10^{-4}$), learning decay (0.999), and batch size (20) were determined with grid search based on average Dice on the validation set.
% A slice is labeled as slice without nodule If there is no segmented nodule in the slice.
\subsubsection{Two-Nodule Detection:} For slices with two nodules, our framework can detect nodules by segmenting the top two activation blobs in the NAM. We tested the detection on a total of 108 slices with two nodules. The 2-GAP model achieves the best detection performance, where both nodules are correctly detected in 50 slices, and one of the two nodules is correctly detected in another 42 slices. With adequate training data, our framework can extend to multi-class classification to automatically determine the number of nodules to segment in the slice.

%%%%%%%%%%%%%%%%%%%%%%%%%%%%%%%%%%%%%%%%%%%%%%%%%%%%%%%%%%%%%%%%%%%%%%%%%%%%%%%%%%%%%%%%%%%%%%%%%%%%
\section{Discussions and Conclusions}

In this work we have proposed an original design for lung nodule segmentation, extending a classification-trained CNN model with GAP operations, to learn discriminative regions at different resolution scales utilizing only weakly labeled training data (present or not of a lung nodule). Coarse-to-fine segmentation extracts nodule candidates using an ICM deformable model, and determines the true nodule exploiting a novel candidate-screening framework. Compared with voxel-based labels, the number of labeling required for our method is reduced by $N_{\mathrm{voxel}}/N_{\mathrm{slice}}\sim100$ times. Detection performance of our weakly-supervised framework compares very favorably with a fully-supervised CNN model (higher TPR and much lower FPR). Our average segmentation accuracy on detected nodules is also very high and gets very close to the benchmark method for larger nodules. Fully-supervised CNN achieves, on average, more accurate segmentation when correctly detecting the nodule, which is expected since voxel-level annotation utilized during training provides more power to deal with various intensity patterns, especially at edges. On the other hand, standard deviations are smaller with the proposed method, hence indicates fewer large mistakes. 

NAM can act as an efficient screening framework that can be incorporated with patch-level labels for false positive reduction \cite{luna16}, or with a small amount of voxel-level labels to learn fine segmentation contour. 
Future work will also extend NAM to 3D CNN to take advantage of the 3D contextual information.
%Future work will also compare different CNN architecture such as ResNet \cite{resnet} and extend NAM to 3D CNN to take advantage of the 3D contextual information.

A machine learning model requiring only weakly-labeled data is key for a sustainable development of CAD systems, as expert time is scarce and expensive and as scanners continue to evolve significantly. Our work used transfer learning from a CNN trained on natural images; with more annotated data, it will be possible to train a fully dedicated network that is likely to be more effective.

\footnotesize{\subsubsection{Acknowledgements:} Thanks NIH R01-HL121270 for funding.}

\bibliographystyle{splncs03_mod}
% \bibliography{paper243_short.bib}

%%%%%%%%%%%%%%%%%%%%%%%%%%%%%%%%%%%%%%%%%%%%%%%%%%%%%%%%%%%%%%%%%%%%%%%%%%%%%%%%%%%%%%%%%%%%%%%%%%%%
\end{document}